\documentclass[conference]{IEEEtran}
\IEEEoverridecommandlockouts

\usepackage{url}
\usepackage{xcolor}
\usepackage{array}
\usepackage{booktabs}
\newcommand{\PreserveBackslash}[1]{\let\temp=\\#1\let\\=\temp}
\newcolumntype{C}[1]{>{\PreserveBackslash\centering}p{#1}}
\newcolumntype{R}[1]{>{\PreserveBackslash\raggedleft}p{#1}}
\newcolumntype{L}[1]{>{\PreserveBackslash\raggedright}p{#1}}

\usepackage{hyperref}
\usepackage{multicol}
\usepackage{multirow}
\begin{document}

\title{Effects of Prompt Length on Domain-specific Tasks for Large Language Models

\author{
\IEEEauthorblockN{Qibang Liu\textsuperscript{1}, Wenzhe Wang\textsuperscript{2}, Jeffrey Willard\textsuperscript{3}}
\IEEEauthorblockA{\textsuperscript{1}Georgia Institute of Technology \textsuperscript{2}Nanjing University of Finance and Economics \textsuperscript{3}Boston University}
\IEEEauthorblockA{qibang@gatech.edu, 2120221363@stu.nufe.edu.cn, jeffwil@bu.edu}

}
}

\maketitle

\begin{abstract}
In recent years, Large Language Models have garnered significant attention for their strong performance in various natural language tasks, such as machine translation and question-answering. These models demonstrate an impressive ability to generalize across diverse tasks. However, their effectiveness in tackling domain-specific tasks, such as financial sentiment analysis and monetary policy understanding, remains a topic of debate, as these tasks often require specialized knowledge and precise reasoning. To address such challenges, researchers design various prompts to unlock the models' abilities. By carefully crafting input prompts, researchers can guide these models to produce more accurate responses. Consequently, prompt engineering has become a key focus of study. Despite the advancements in both models and prompt engineering, the relationship between the two---specifically, how prompt design impacts models' ability to perform domain-specific tasks---remains underexplored. This paper aims to bridge this research gap.
\end{abstract}


\section{Introduction}

Large Language Models (LLMs) have become an integral part of our daily lives, transforming the way we interact with technology. 
From virtual assistants like Siri and Alexa to language translation apps like Google Translate, LLMs are being used in various applications to make our lives easier. 
For instance, chatbots powered by LLMs are being used by businesses to provide customer support and answer frequently asked questions. 
Additionally, LLMs are being used in writing assistants like Grammarly to help users improve their writing skills. 
Furthermore, LLMs are also being used in language learning apps like Duolingo to help users learn new languages. 

Prompts play a crucial role in unlocking the potential of LLMs. 
A well-crafted prompt can guide the LLM to generate accurate, relevant, and coherent responses. 
For instance, when asking an LLM to write a story, providing a prompt with specific details such as genre, setting, and characters can help the model generate a more engaging and focused story. 
However, despite of the well-known advantages, LLMs' performance relies heavily on the quality and specificity of the prompts. 
Poorly constructed prompts can lead to inaccurate outputs because LLMs lack intrinsic knowledge of research. 
For example, LLMs may output oversimplified or non-viable solutions due to not inherently understanding the research mechanics. 
This limitation emphasizes the domain expertise in crafting prompts and interpreting results.

Through diversifying prompt instructions, researchers pioneeringly address the LLMs' performance across specific domains (e.g., economics, health, and technical systems), which empowers nations to strengthen economic resilience, protect public health, and maintain technological leadership on a global scale. By systematically examining domain-specific impacts, their works highlight the transformative potential of prompt engineering in identifying LLM capabilities for specialized applications. For example, LLMs help with policy-making by understanding financial sentiment \cite{10.1145/3696271.3696299}. In public health, LLMs revolutionize disease detection, enabling personalized treatments and providing health support \cite{wei2024enhancingdiseasedetectionradiology}. 
vestigation.

While various aspects of prompts have been extensively researched, such as prompt design, and their content, one crucial aspect remains unexplored: the length of prompts. 
Despite the growing importance of LLMs, the optimal length of prompts that can effectively elicit desired responses from these models is still unknown. 
Previous research has focused on optimizing prompt content, structure, and style to improve model performance, but the impact of prompt length on model behavior and response quality has not been systematically investigated. 
As a result, we aim to bridge this gap in the understanding of how prompt length affects LLM performance in domain-specific tasks.

Our main contributions are as follows. 
\begin{itemize}
    \item We conduct extensive experiments with LLMs on nine domain-specific tasks, and the results show that although they claim to be state-of-the-art NLP models at the time of their releases, with the default prompt length, they still struggle to tackle tasks without sufficient domain knowledge.
    \item Our results with different prompt lengths show that long prompts, providing more background knowledge about the domain that the task falls into, are generally beneficial.
    \item Our results also show that even with background knowledge in long prompts, LLMs' performance still lags behind humans, as the average F1-score is much less than 1.0.
\end{itemize}

\section{Background}

\subsection{Deep Learning and Large Language Model}

Since the introduction of AlexNet~\cite{NIPS2012_c399862d}, we have entered the era of deep learning, marking significant advancements across various fields, such as asphalt pavement analysis~\cite{DAN2024134837,Dan31122024}, scene text detection~\cite{Tang_2022_CVPR,10239535}, visual question answering~\cite{tang2024mtvqabenchmarkingmultilingualtextcentric,tang2024textsquarescalingtextcentricvisual}, scene text recognition~\cite{10.1145/3503161.3547787,10.1007/978-3-031-19815-1_14,Zhao_2024_CVPR}, character recognition~\cite{10.1093/nsr/nwad141}, and adversarial attach~\cite{10.1117/12.3009240,10263390}. In addition to the application layer, theoretical investigation has attracted substantial attention, such as Q Learning~\cite{zhao2024minimaxoptimalqlearning}.
In NLP, the progress in language model pretraining began with Word2Vec~\cite{NIPS2013_9aa42b31,Mikolov2013EfficientEO} and has accelerated rapidly. Among these developments, the Transformer architecture~\cite{NIPS2017_3f5ee243} is the backbone of all modern language models. 

Transformer's two core components, the encoder and decoder, have been the foundation for much of the work in model pretraining. For encoder-based pretraining, BERT~\cite{devlin-etal-2019-bert} is the most representative example, inspiring numerous variants to improve the pretraining paradigm (e.g., RoBERTa~\cite{liu2019robertarobustlyoptimizedbert}, DistillBERT~\cite{Sanh2019DistilBERTAD}, and ALBERT~\cite{DBLP:conf/iclr/LanCGGSS20} to address domain-specific tasks (e.g., FinBERT~\cite{10.5555/3491440.3492062} for finance, SciBERT~\cite{beltagy-etal-2019-scibert} for scientific texts, and ClinicalBERT~\cite{clinicalbert} for medical data).
Simultaneously, another branch of pretraining work focuses on the decoder of the Transformer. The most notable examples include the GPT series (e.g., GPT-1~\cite{Radford2018ImprovingLU}, GPT-2~\cite{Radford2019LanguageMA}, GPT-3~\cite{NEURIPS2020_1457c0d6}) and the LLaMA series (e.g., LLaMA-1~\cite{touvron2023llamaopenefficientfoundation}, LLaMA-2~\cite{touvron2023llama2openfoundation}, LLaMA-3~\cite{grattafiori2024llama3herdmodels}, and its incremental updates like LLaMA 3.1, 3.2, and 3.3\footnote{\url{https://www.llama.com/docs/model-cards-and-prompt-formats}}). These models discard the encoder and rely solely on the decoder as their backbone architecture. 
Following the rise of ChatGPT\footnote{chatgpt.com}, which fueled widespread interest in LLMs, researchers have increasingly focused on evaluating the LLaMA series on various domain-specific tasks, such as financial sentiment analysis~\cite{10.1145/3696271.3696299} and emotion identification~\cite{10.1145/3696271.3696292}.

However, no prior work has investigated how prompt length impacts the LLM performance on domain-specific tasks. This is important because earlier studies have shown that prompts are crucial in eliciting LLMs' abilities for effective language understanding and task performance~\cite{10458651,xiao-etal-2024-analyzing,10.1145/3655497.3655500}. This paper aims to provide insights to bridge this gap.

\subsection{Prompt Engineering}

Most existing work on prompt engineering focuses on enhancing reasoning and logical capabilities in language models. Specifically, Chain-of-Thought (CoT) prompting~\cite{10.5555/3600270.3602070,zhang2022automaticchainthoughtprompting}, which systematically investigates step-by-step reasoning, is considered a pioneering approach in this domain. 
Building on CoT, subsequent advancements have been proposed: Self-consistency~\cite{wang2023selfconsistencyimproveschainthought} introduces mechanisms to improve reliability in reasoning paths; Logical CoT prompting~\cite{zhao-etal-2024-enhancing-zero} refines logical reasoning within prompts; Chain-of-Symbols (CoS) prompting~\cite{hu2024chainofsymbolpromptingelicitsplanning} explores symbolic representations to enhance task-solving processes; Tree-of-Thoughts prompting~\cite{10.5555/3666122.3666639} introduces hierarchical reasoning structures; Graph-of-Thoughts (GoT) prompting~\cite{yao-etal-2024-got} leverages interconnected reasoning nodes to expand context understanding; Thread-of-Thought (ThouT) prompting~\cite{zhou2023threadthoughtunravelingchaotic} emphasizes linear and contextual reasoning threads; and Chain-of-Tables prompting~\cite{wang2024chainoftableevolvingtablesreasoning} adapts tabular representations for specific reasoning tasks.

Although these innovations advance the capabilities of language models in handling complex reasoning and logic-intensive applications, there is not a comprehensive study of how the length of prompt affects LLMs' capability in domain-specific tasks.

\section{Experiments}

\begin{table}[t]
    \centering
    \small
    \begin{tabular}{l|l}
        \toprule
        Acronym & Full name \\
        \midrule
        MPU & Monetary Policy Understanding \\ 
        UI & User Intent \\  
        CD & Conversation Domain \\
        QIC & Query Intent Classification \\        
        SD & Sarcasm Detection \\
        EI & Emotion Identification \\         
        FSA & Financial Sentiment Analysis \\        
        TSB & Technical System Behavior \\
        DD & Disease Detection \\

        \bottomrule
        \addlinespace
    \end{tabular}
    \caption{Acronyms of the domain-specific tasks in this paper.}
    \label{tab:task_name}
\end{table}

To assess the impact of prompt length on LLMs' performance across domain-specific tasks, we conducted a series of comprehensive experiments under varying prompt length settings. Specifically, we conducted nine groups of experiments: Monetary Policy Understanding, User Intent, Conversation Domain, Query Intent Classification, Sarcasm Detection, Emotion Identification, Financial Sentiment Analysis, Technical System Behavior, and Disease Detection.  
We use their acronyms throughout this experimental section and the subsequent results section for clarity and improved readability. Table~\ref{tab:task_name} maps the task acronyms and their full names.

\begin{table*}
    \hfill
    \begin{minipage}[l]{0.47\textwidth}
        \centering
    \small
    \begin{tabular}{l|r|r|r|}
        \toprule
         & \multicolumn{1}{c|}{Precision} & \multicolumn{1}{c|}{Recall} & \multicolumn{1}{c|}{F1 Score} \\ 
        \midrule
        
        MPU\textsubscript{base}~\cite{10871796} & 0.53 & 0.52 & 0.51  \\
        MPU\textsubscript{base} + LI & \textcolor{teal}{+(0.02)} \textcolor{teal}{0.55} & \textcolor{teal}{+(0.03)} \textcolor{teal}{0.55} & \textcolor{teal}{+(0.04)} \textcolor{teal}{0.55} \\
        MPU\textsubscript{base} + SI & \textcolor{red}{-(0.03)} \textcolor{red}{0.50} & \textcolor{red}{-(0.07)} \textcolor{red}{0.45} & \textcolor{red}{-(0.04)} \textcolor{red}{0.47} \\
        
        \midrule
        
        UI\textsubscript{base} \cite{10.1145/3637528.3671647} & 0.52 & 0.74 & 0.61 \\ 
        UI\textsubscript{base} + LI & \textcolor{teal}{+(0.04)} \textcolor{teal}{0.56} & \textcolor{teal}{+(0.02)} \textcolor{teal}{0.76} & \textcolor{teal}{+(0.02)} \textcolor{teal}{0.63} \\
        UI\textsubscript{base} + SI & \textcolor{red}{-(0.03)} \textcolor{red}{0.49} & \textcolor{red}{-(0.04)} \textcolor{red}{0.70} & \textcolor{red}{-(0.03)} \textcolor{red}{0.58} \\
        
        \midrule
        
        CD\textsubscript{base} \cite{10.1145/3637528.3671647} & 0.44 & 0.75 & 0.56 \\ 
        CD\textsubscript{base} + LI& \textcolor{black}{(0.00)} \textcolor{black}{0.44} & \textcolor{teal}{+(0.01)} \textcolor{teal}{0.76} & \textcolor{teal}{+(0.01)} \textcolor{teal}{0.57} \\ 
        CD\textsubscript{base} + SI & \textcolor{red}{-(0.01)} \textcolor{red}{0.43} & \textcolor{red}{-(0.01)} \textcolor{red}{0.74} & \textcolor{red}{-(0.01)} \textcolor{red}{0.55} \\ 
        
        \midrule
        
        QIC\textsubscript{base} \cite{javadi2024llmbasedweaksupervisionframework} & 0.48 & 0.10 & 0.84 \\ 
        QIC\textsubscript{base} + LI & \textcolor{teal}{+(0.06)} \textcolor{teal}{0.54} & \textcolor{teal}{+(0.10)} \textcolor{teal}{0.20} & \textcolor{teal}{+(0.08)} \textcolor{teal}{0.92} \\ 
        QIC\textsubscript{base} + SI & \textcolor{red}{-(0.06)} \textcolor{red}{0.42} & \textcolor{red}{-(0.02)} \textcolor{red}{0.08} & \textcolor{red}{-(0.04)} \textcolor{red}{0.80} \\

        \midrule
        
    SD\textsubscript{base} \cite{10.1145/3696271.3696294} & 0.67 & 0.66 & 0.66 \\
        SD\textsubscript{base} + LI & \textcolor{teal}{+(0.06)} \textcolor{teal}{0.73} & \textcolor{teal}{+(0.04)} \textcolor{teal}{0.70} & \textcolor{teal}{+(0.05)} \textcolor{teal}{0.71} \\
        SD\textsubscript{base} + SI & \textcolor{red}{-(0.03)} \textcolor{red}{0.64} & \textcolor{red}{-(0.01)} \textcolor{red}{0.65} & \textcolor{red}{-(0.02)} \textcolor{red}{0.64} \\
        
        \bottomrule
        \addlinespace
        
    \end{tabular}
    \end{minipage}
    \hspace{-1.0cm}
    \hfill
    \begin{minipage}[l]{0.47\textwidth}
        \centering
    \small
    \begin{tabular}{l|r|r|r}
        \toprule
         & \multicolumn{1}{c|}{Precision} & \multicolumn{1}{c|}{Recall} & \multicolumn{1}{c}{F1 Score} \\ 
        \midrule
        
        EI\textsubscript{base} \cite{10.1145/3696271.3696292} & 0.49 & 0.50 & 0.48 \\
        EI\textsubscript{base} + LI & \textcolor{teal}{+(0.01)} \textcolor{teal}{0.50} & \textcolor{black}{(0.00)} \textcolor{teal}{0.50} & \textcolor{teal}{+(0.01)} \textcolor{teal}{0.49} \\
        EI\textsubscript{base} + SI & \textcolor{red}{-(0.04)} \textcolor{red}{0.45} & \textcolor{red}{-(0.06)} \textcolor{red}{0.44} & \textcolor{red}{-(0.05)} \textcolor{red}{0.43} \\
        
        \midrule
        
        FSA\textsubscript{base} \cite{10.1145/3696271.3696299} & 0.66 & 0.75 & 0.67\\
        FSA\textsubscript{base} + LI & \textcolor{teal}{+(0.02)} \textcolor{teal}{0.68} & \textcolor{teal}{+(0.08)} \textcolor{teal}{0.83} & \textcolor{teal}{+(0.05)} \textcolor{teal}{0.72} \\
        FSA\textsubscript{base} + SI & \textcolor{red}{-(0.05)} \textcolor{red}{0.61} & \textcolor{red}{-(0.08)} \textcolor{red}{0.67} & \textcolor{red}{-(0.07)} \textcolor{red}{0.60} \\

        \midrule
        
        
        TSB\textsubscript{base} \cite{zhang2024empowering} & 0.36 & 0.49 & 0.42 \\
        TSB\textsubscript{base} + LI & \textcolor{teal}{+(0.05)} \textcolor{teal}{0.41} & \textcolor{teal}{+(0.09)} \textcolor{teal}{0.58} & \textcolor{teal}{+(0.07)} \textcolor{teal}{0.49} \\
        TSB\textsubscript{base} + SI & \textcolor{red}{-(0.02)} \textcolor{red}{0.34} & \textcolor{red}{-(0.05)} \textcolor{red}{0.44} & \textcolor{red}{-(0.03)} \textcolor{red}{0.39} \\

        \midrule
        DD\textsubscript{base} \cite{wei2024enhancingdiseasedetectionradiology} & 0.83 & 0.89 & 0.86 \\
        DD\textsubscript{base} + LI & \textcolor{teal}{+(0.01)} \textcolor{teal}{0.84} & \textcolor{teal}{+(0.01)} \textcolor{teal}{0.90} & \textcolor{teal}{+(0.01)} \textcolor{teal}{0.87} \\
        DD\textsubscript{base} + SI & \textcolor{red}{-(0.08)} \textcolor{red}{0.74} & \textcolor{red}{-(0.12)} \textcolor{red}{0.77} & \textcolor{red}{-(0.09)} \textcolor{red}{0.77} \\
        \midrule

        \multicolumn{4}{l}{} \\
        \multicolumn{4}{l}{\textbf{Note:} LI = Long Instructions, SI = Short Instructions} \\
        \multicolumn{4}{l}{} \\


         \bottomrule
        \addlinespace
    \end{tabular}
    \end{minipage}
         
         
    \caption{Experimental results on domain specific tasks with different prompt lengths. }
    \label{table:results}
\end{table*}

\subsection{Domain-specific Tasks}

MPU aims to classify monetary policy statements as hawkish, dovish, or neutral, where hawkish signals tightening measures like higher interest rates, dovish suggests easing policies to stimulate the economy, and neutral reflects no significant stance. 
UI classifies the users' intent based on what they say.
CD categorizes conversations into specific domains or topics, such as healthcare, technology, or finance, for more accurate contextual analysis. 
QIC determines the underlying intent behind user queries, such as whether the query seeks information, navigational guidance, or specific documents, enabling more efficient document retrieval systems. 
SD focuses on the sarcasm that may potentially exist in people's words.
EI detects specific emotions---anger, joy, sadness, surprise, fear, or love---expressed in a given sentence to better understand emotional tone and context.
FSA aims to evaluate financial texts to determine whether the sentiment conveyed is positive, negative, or neutral, helping assess market outlooks or economic conditions. 
SD focuses on identifying whether a statement contains sarcasm, which is crucial for understanding the true intent of a speaker or writer. 
TSB studies the technical performances, such as end-to-end delay and throughput, which are significant to domain services such as online video and autonomous driving.
Finally, DD aims to extract abnormal findings from radiology reports corresponding to ICD-10 codes.

\subsection{Experiment Design}

In our experiment, each group consists of three settings: default prompt length\footnote{The default prompt length can be viewed as ``mid length''.}, short instruction, and long instruction. 
The default prompts that we experimented with are the prompts provided in the original papers.
We define short instructions as those containing less than 50\% tokens of the default prompt, typically only describing the task name. 
In contrast, long instructions contain at least 200\% tokens of the default prompt, providing not only requirements but also background knowledge and experimental conditions that can help in answering questions. 
For TSB, while the prompts are not explicitly provided in the original paper, the default instructions cover all the conditions that were introduced in the referenced work, such as hardware model, software version, and experimental environments. The rationale behind choosing these base works is their popularity in their domains.
To the best of our knowledge, these works are the most recent works leveraging LLMs in their domains.
This design allows us to investigate how different prompt lengths impact LLMs' performance and ability to leverage contextual information.

To evaluate our experiment results, we define a correct case for each scenario. For classification problems, such as MPU, FSA, and SD, a case is considered correct if LLMs produce the same decision as the ground truth. Regarding the engineered system where the prediction may cover a range, we divide the possible range (e.g., $[0,100]$) into ten equal segments, offering a structured framework to evaluate the results compared to the ground truth. We treat segments fairly without considering them reliable or reasonable, as our goal is to ensure an unbiased performance assessment \cite{URLLC}. 

We use the weighted average precision, recall, and F1-score for each experiment as the evaluation metrics, where the weight is determined by the number of instances for each class. 
Precision measures how many positive predictions are correct (True Positive, TP for short. False Positive, FP for short.). 
Recall is a measure of how many of the positive cases the classifier correctly predicted over all the positive cases in the test data (False Negative, FN for short).
F1 score is a measure combining both precision and recall.

\section{Experimental Results}

Based on the results presented in Table~\ref{table:results}, we observe distinct trends in how different instruction strategies impact the performance of precision, recall, and F1-score across domain-specific tasks.\footnote{To ensure the generalizability of our results, we run each experiment 10 times under the same experiment setting, and report the average results in Table~\ref{table:results}.}

\noindent \textbf{Results of default prompt length.} The baseline performance metrics vary significantly across tasks. The domains of SD, FSA, and TSB exhibit strong baseline performance, with F1-scores around 0.67.
The domain of QIC shows high F1-score (0.84) but low recall (0.10), indicating potential issues with sensitivity to positive cases, while the domain of DD is excellent with precision, recall, and F1-score above 0.80. The domains of EI and CD exhibit relatively lower baseline performance, with F1-scores of 0.48 and 0.56, respectively.

\noindent \textbf{Results of short prompts.} Short instructions negatively affect performances in all tasks compared to baseline performance. In QIC and DD domains, performances significantly decrease, with precision dropping by 0.06 and 0.08, recall by 0.02 and 0.12, and F1-score by 0.04 and 0.09, while CD maintains slight drops with precision, recall, and F1-score all by 0.01. These results suggest that short instructions can not fully leverage LLMs' capabilities, particularly in sufficient details-needed tasks where contextual information (QIC) or special fields (DD) is essential.

\noindent \textbf{Results of long prompts.} It is worth noting that long instructions generally improve the performance metrics across all tasks on all experimented domains compared to base performance. Specifically, in EI and DD, minimal improvements are observed, with precision and F1-score increasing by +0.01 each, while QIC and TSB show the biggest improved performances due to their detail-sensitive tasks. 
Moreover, even with the detailed background knowledge provided in the long prompt, LLMs still struggle in these tasks, as their F1 scores are far behind 1.0 (which represents human-level understanding ability).

\section{Conclusions}

In this study, we conducted comprehensive experiments to assess the effect of prompt length on LLMs' performance in various domain-specific tasks. Our findings indicate that longer prompts generally enhance model performance, while shorter prompts can be detrimental. Furthermore, despite providing extensive background knowledge of the prompts, LLMs still struggle with challenging domain-specific tasks, highlighting the need for a deeper understanding of prompt phrasing.

Our research agenda will be centered on exploring how different prompting techniques influence LLMs' performance in domain-specific tasks. Recent studies have shown that modifying the phrasing of questions in prompts and adjusting the number of examples provided in instructions can significantly improve performance in specialized domains, including spatial information extraction~\cite{xiao-etal-2024-analyzing}, math reasoning~\cite{10.5555/3600270.3602070, yao-etal-2024-got}, short interest understanding~\cite{10.1145/3655497.3655500}, and corporate event prediction~\cite{10458651}. These findings highlight the potential benefits of optimizing prompting techniques to enhance LLMs' performance in specific tasks.

\bibliography{cm1}

\begin{thebibliography}{10}
\providecommand{\url}[1]{#1}
\csname url@samestyle\endcsname
\providecommand{\newblock}{\relax}
\providecommand{\bibinfo}[2]{#2}
\providecommand{\BIBentrySTDinterwordspacing}{\spaceskip=0pt\relax}
\providecommand{\BIBentryALTinterwordstretchfactor}{4}
\providecommand{\BIBentryALTinterwordspacing}{\spaceskip=\fontdimen2\font plus
\BIBentryALTinterwordstretchfactor\fontdimen3\font minus \fontdimen4\font\relax}
\providecommand{\BIBforeignlanguage}[2]{{%
\expandafter\ifx\csname l@#1\endcsname\relax
\typeout{** WARNING: IEEEtran.bst: No hyphenation pattern has been}%
\typeout{** loaded for the language `#1'. Using the pattern for}%
\typeout{** the default language instead.}%
\else
\language=\csname l@#1\endcsname
\fi
#2}}
\providecommand{\BIBdecl}{\relax}
\BIBdecl

\bibitem{10.1145/3696271.3696299}
\BIBentryALTinterwordspacing
Z.~Mai, J.~Zhang, Z.~Xu, and Z.~Xiao, ``Financial sentiment analysis meets llama 3: A comprehensive analysis,'' in \emph{Proceedings of the 2024 7th International Conference on Machine Learning and Machine Intelligence (MLMI)}, ser. MLMI '24.\hskip 1em plus 0.5em minus 0.4em\relax New York, NY, USA: Association for Computing Machinery, 2024, p. 171–175. [Online]. Available: \url{https://doi.org/10.1145/3696271.3696299}
\BIBentrySTDinterwordspacing

\bibitem{wei2024enhancingdiseasedetectionradiology}
\BIBentryALTinterwordspacing
Y.~Wei, X.~Wang, H.~Ong, Y.~Zhou, A.~Flanders, G.~Shih, and Y.~Peng, ``Enhancing disease detection in radiology reports through fine-tuning lightweight llm on weak labels,'' 2024. [Online]. Available: \url{https://arxiv.org/abs/2409.16563}
\BIBentrySTDinterwordspacing

\bibitem{NIPS2012_c399862d}
\BIBentryALTinterwordspacing
A.~Krizhevsky, I.~Sutskever, and G.~E. Hinton, ``Imagenet classification with deep convolutional neural networks,'' in \emph{Advances in Neural Information Processing Systems}, F.~Pereira, C.~Burges, L.~Bottou, and K.~Weinberger, Eds., vol.~25.\hskip 1em plus 0.5em minus 0.4em\relax Curran Associates, Inc., 2012. [Online]. Available: \url{https://proceedings.neurips.cc/paper_files/paper/2012/file/c399862d3b9d6b76c8436e924a68c45b-Paper.pdf}
\BIBentrySTDinterwordspacing

\bibitem{DAN2024134837}
\BIBentryALTinterwordspacing
H.-C. Dan, B.~Lu, and M.~Li, ``Evaluation of asphalt pavement texture using multiview stereo reconstruction based on deep learning,'' \emph{Construction and Building Materials}, vol. 412, p. 134837, 2024. [Online]. Available: \url{https://www.sciencedirect.com/science/article/pii/S0950061823045580}
\BIBentrySTDinterwordspacing

\bibitem{Dan31122024}
H.-C. Dan, P.~Yan, J.~Tan, Y.~Zhou, and B.~Lu, ``Multiple distresses detection for asphalt pavement using improved you only look once algorithm based on convolutional neural network,'' \emph{International Journal of Pavement Engineering}, vol.~25, no.~1, p. 2308169, 2024.

\bibitem{Tang_2022_CVPR}
J.~Tang, W.~Zhang, H.~Liu, M.~Yang, B.~Jiang, G.~Hu, and X.~Bai, ``Few could be better than all: Feature sampling and grouping for scene text detection,'' in \emph{Proceedings of the IEEE/CVF Conference on Computer Vision and Pattern Recognition (CVPR)}, June 2022, pp. 4563--4572.

\bibitem{10239535}
Y.~Liu, J.~Zhang, D.~Peng, M.~Huang, X.~Wang, J.~Tang, C.~Huang, D.~Lin, C.~Shen, X.~Bai, and L.~Jin, ``Spts v2: Single-point scene text spotting,'' \emph{IEEE Transactions on Pattern Analysis and Machine Intelligence}, vol.~45, no.~12, pp. 15\,665--15\,679, 2023.

\bibitem{tang2024mtvqabenchmarkingmultilingualtextcentric}
\BIBentryALTinterwordspacing
J.~Tang, Q.~Liu, Y.~Ye, J.~Lu, S.~Wei, C.~Lin, W.~Li, M.~F. F.~B. Mahmood, H.~Feng, Z.~Zhao, Y.~Wang, Y.~Liu, H.~Liu, X.~Bai, and C.~Huang, ``Mtvqa: Benchmarking multilingual text-centric visual question answering,'' 2024. [Online]. Available: \url{https://arxiv.org/abs/2405.11985}
\BIBentrySTDinterwordspacing

\bibitem{tang2024textsquarescalingtextcentricvisual}
\BIBentryALTinterwordspacing
J.~Tang, C.~Lin, Z.~Zhao, S.~Wei, B.~Wu, Q.~Liu, H.~Feng, Y.~Li, S.~Wang, L.~Liao, W.~Shi, Y.~Liu, H.~Liu, Y.~Xie, X.~Bai, and C.~Huang, ``Textsquare: Scaling up text-centric visual instruction tuning,'' 2024. [Online]. Available: \url{https://arxiv.org/abs/2404.12803}
\BIBentrySTDinterwordspacing

\bibitem{10.1145/3503161.3547787}
\BIBentryALTinterwordspacing
J.~Tang, S.~Qiao, B.~Cui, Y.~Ma, S.~Zhang, and D.~Kanoulas, ``You can even annotate text with voice: Transcription-only-supervised text spotting,'' in \emph{Proceedings of the 30th ACM International Conference on Multimedia}, ser. MM '22.\hskip 1em plus 0.5em minus 0.4em\relax New York, NY, USA: Association for Computing Machinery, 2022, p. 4154–4163. [Online]. Available: \url{https://doi.org/10.1145/3503161.3547787}
\BIBentrySTDinterwordspacing

\bibitem{10.1007/978-3-031-19815-1_14}
J.~Tang, W.~Qian, L.~Song, X.~Dong, L.~Li, and X.~Bai, ``Optimal boxes: Boosting end-to-end scene text recognition by adjusting annotated bounding boxes via reinforcement learning,'' in \emph{Computer Vision -- ECCV 2022}, S.~Avidan, G.~Brostow, M.~Ciss{\'e}, G.~M. Farinella, and T.~Hassner, Eds.\hskip 1em plus 0.5em minus 0.4em\relax Cham: Springer Nature Switzerland, 2022, pp. 233--248.

\bibitem{Zhao_2024_CVPR}
Z.~Zhao, J.~Tang, C.~Lin, B.~Wu, C.~Huang, H.~Liu, X.~Tan, Z.~Zhang, and Y.~Xie, ``Multi-modal in-context learning makes an ego-evolving scene text recognizer,'' in \emph{Proceedings of the IEEE/CVF Conference on Computer Vision and Pattern Recognition (CVPR)}, June 2024, pp. 15\,567--15\,576.

\bibitem{10.1093/nsr/nwad141}
\BIBentryALTinterwordspacing
J.~Tang, W.~Du, B.~Wang, W.~Zhou, S.~Mei, T.~Xue, X.~Xu, and H.~Zhang, ``Character recognition competition for street view shop signs,'' \emph{National Science Review}, vol.~10, no.~6, p. nwad141, 05 2023. [Online]. Available: \url{https://doi.org/10.1093/nsr/nwad141}
\BIBentrySTDinterwordspacing

\bibitem{10.1117/12.3009240}
\BIBentryALTinterwordspacing
Y.~Tao, ``{SQBA: sequential query-based blackbox attack},'' in \emph{Fifth International Conference on Artificial Intelligence and Computer Science (AICS 2023)}, H.~Zaidi, Y.~S. Shmaliy, H.~Meng, H.~Kolivand, Y.~Sun, J.~Luo, and M.~Alazab, Eds., vol. 12803, International Society for Optics and Photonics.\hskip 1em plus 0.5em minus 0.4em\relax SPIE, 2023, p. 128032Q. [Online]. Available: \url{https://doi.org/10.1117/12.3009240}
\BIBentrySTDinterwordspacing

\bibitem{10263390}
------, ``Meta learning enabled adversarial defense,'' in \emph{2023 IEEE International Conference on Sensors, Electronics and Computer Engineering (ICSECE)}, 2023, pp. 1326--1330.

\bibitem{zhao2024minimaxoptimalqlearning}
\BIBentryALTinterwordspacing
P.~Zhao and L.~Lai, ``Minimax optimal q learning with nearest neighbors,'' 2024. [Online]. Available: \url{https://arxiv.org/abs/2308.01490}
\BIBentrySTDinterwordspacing

\bibitem{NIPS2013_9aa42b31}
\BIBentryALTinterwordspacing
T.~Mikolov, I.~Sutskever, K.~Chen, G.~S. Corrado, and J.~Dean, ``Distributed representations of words and phrases and their compositionality,'' in \emph{Advances in Neural Information Processing Systems}, C.~Burges, L.~Bottou, M.~Welling, Z.~Ghahramani, and K.~Weinberger, Eds., vol.~26.\hskip 1em plus 0.5em minus 0.4em\relax Curran Associates, Inc., 2013. [Online]. Available: \url{https://proceedings.neurips.cc/paper_files/paper/2013/file/9aa42b31882ec039965f3c4923ce901b-Paper.pdf}
\BIBentrySTDinterwordspacing

\bibitem{Mikolov2013EfficientEO}
\BIBentryALTinterwordspacing
T.~Mikolov, K.~Chen, G.~Corrado, and J.~Dean, ``Efficient estimation of word representations in vector space,'' in \emph{1st International Conference on Learning Representations, {ICLR} 2013, Scottsdale, Arizona, USA, May 2-4, 2013, Workshop Track Proceedings}, Y.~Bengio and Y.~LeCun, Eds., 2013. [Online]. Available: \url{http://arxiv.org/abs/1301.3781}
\BIBentrySTDinterwordspacing

\bibitem{NIPS2017_3f5ee243}
\BIBentryALTinterwordspacing
A.~Vaswani, N.~Shazeer, N.~Parmar, J.~Uszkoreit, L.~Jones, A.~N. Gomez, L.~u. Kaiser, and I.~Polosukhin, ``Attention is all you need,'' in \emph{Advances in Neural Information Processing Systems}, I.~Guyon, U.~V. Luxburg, S.~Bengio, H.~Wallach, R.~Fergus, S.~Vishwanathan, and R.~Garnett, Eds., vol.~30.\hskip 1em plus 0.5em minus 0.4em\relax Curran Associates, Inc., 2017. [Online]. Available: \url{https://proceedings.neurips.cc/paper_files/paper/2017/file/3f5ee243547dee91fbd053c1c4a845aa-Paper.pdf}
\BIBentrySTDinterwordspacing

\bibitem{devlin-etal-2019-bert}
\BIBentryALTinterwordspacing
J.~Devlin, M.-W. Chang, K.~Lee, and K.~Toutanova, ``{BERT}: Pre-training of deep bidirectional transformers for language understanding,'' in \emph{Proceedings of the 2019 Conference of the North {A}merican Chapter of the Association for Computational Linguistics: Human Language Technologies, Volume 1 (Long and Short Papers)}, J.~Burstein, C.~Doran, and T.~Solorio, Eds.\hskip 1em plus 0.5em minus 0.4em\relax Minneapolis, Minnesota: Association for Computational Linguistics, Jun. 2019, pp. 4171--4186. [Online]. Available: \url{https://aclanthology.org/N19-1423}
\BIBentrySTDinterwordspacing

\bibitem{liu2019robertarobustlyoptimizedbert}
\BIBentryALTinterwordspacing
Y.~Liu, M.~Ott, N.~Goyal, J.~Du, M.~Joshi, D.~Chen, O.~Levy, M.~Lewis, L.~Zettlemoyer, and V.~Stoyanov, ``Roberta: A robustly optimized bert pretraining approach,'' 2019. [Online]. Available: \url{https://arxiv.org/abs/1907.11692}
\BIBentrySTDinterwordspacing

\bibitem{Sanh2019DistilBERTAD}
\BIBentryALTinterwordspacing
V.~Sanh, L.~Debut, J.~Chaumond, and T.~Wolf, ``Distilbert, a distilled version of bert: smaller, faster, cheaper and lighter,'' \emph{ArXiv}, vol. abs/1910.01108, 2019. [Online]. Available: \url{https://api.semanticscholar.org/CorpusID:203626972}
\BIBentrySTDinterwordspacing

\bibitem{DBLP:conf/iclr/LanCGGSS20}
\BIBentryALTinterwordspacing
Z.~Lan, M.~Chen, S.~Goodman, K.~Gimpel, P.~Sharma, and R.~Soricut, ``{ALBERT:} {A} lite {BERT} for self-supervised learning of language representations,'' in \emph{8th International Conference on Learning Representations, {ICLR} 2020, Addis Ababa, Ethiopia, April 26-30, 2020}.\hskip 1em plus 0.5em minus 0.4em\relax OpenReview.net, 2020. [Online]. Available: \url{https://openreview.net/forum?id=H1eA7AEtvS}
\BIBentrySTDinterwordspacing

\bibitem{10.5555/3491440.3492062}
Z.~Liu, D.~Huang, K.~Huang, Z.~Li, and J.~Zhao, ``Finbert: a pre-trained financial language representation model for financial text mining,'' in \emph{Proceedings of the Twenty-Ninth International Joint Conference on Artificial Intelligence}, ser. IJCAI'20, 2021.

\bibitem{beltagy-etal-2019-scibert}
\BIBentryALTinterwordspacing
I.~Beltagy, K.~Lo, and A.~Cohan, ``{S}ci{BERT}: A pretrained language model for scientific text,'' in \emph{Proceedings of the 2019 Conference on Empirical Methods in Natural Language Processing and the 9th International Joint Conference on Natural Language Processing (EMNLP-IJCNLP)}, K.~Inui, J.~Jiang, V.~Ng, and X.~Wan, Eds.\hskip 1em plus 0.5em minus 0.4em\relax Hong Kong, China: Association for Computational Linguistics, Nov. 2019, pp. 3615--3620. [Online]. Available: \url{https://aclanthology.org/D19-1371}
\BIBentrySTDinterwordspacing

\bibitem{clinicalbert}
K.~Huang, J.~Altosaar, and R.~Ranganath, ``Clinicalbert: Modeling clinical notes and predicting hospital readmission,'' 2019.

\bibitem{Radford2018ImprovingLU}
\BIBentryALTinterwordspacing
A.~Radford and K.~Narasimhan, ``Improving language understanding by generative pre-training,'' 2018. [Online]. Available: \url{https://api.semanticscholar.org/CorpusID:49313245}
\BIBentrySTDinterwordspacing

\bibitem{Radford2019LanguageMA}
\BIBentryALTinterwordspacing
A.~Radford, J.~Wu, R.~Child, D.~Luan, D.~Amodei, and I.~Sutskever, ``Language models are unsupervised multitask learners,'' 2019. [Online]. Available: \url{https://api.semanticscholar.org/CorpusID:160025533}
\BIBentrySTDinterwordspacing

\bibitem{NEURIPS2020_1457c0d6}
\BIBentryALTinterwordspacing
T.~Brown, B.~Mann, N.~Ryder, M.~Subbiah, J.~D. Kaplan, P.~Dhariwal, A.~Neelakantan, P.~Shyam, G.~Sastry, A.~Askell, S.~Agarwal, A.~Herbert-Voss, G.~Krueger, T.~Henighan, R.~Child, A.~Ramesh, D.~Ziegler, J.~Wu, C.~Winter, C.~Hesse, M.~Chen, E.~Sigler, M.~Litwin, S.~Gray, B.~Chess, J.~Clark, C.~Berner, S.~McCandlish, A.~Radford, I.~Sutskever, and D.~Amodei, ``Language models are few-shot learners,'' in \emph{Advances in Neural Information Processing Systems}, H.~Larochelle, M.~Ranzato, R.~Hadsell, M.~Balcan, and H.~Lin, Eds., vol.~33.\hskip 1em plus 0.5em minus 0.4em\relax Curran Associates, Inc., 2020, pp. 1877--1901. [Online]. Available: \url{https://proceedings.neurips.cc/paper_files/paper/2020/file/1457c0d6bfcb4967418bfb8ac142f64a-Paper.pdf}
\BIBentrySTDinterwordspacing

\bibitem{touvron2023llamaopenefficientfoundation}
\BIBentryALTinterwordspacing
H.~Touvron, T.~Lavril, G.~Izacard, X.~Martinet, M.-A. Lachaux, T.~Lacroix, B.~Rozière, N.~Goyal, E.~Hambro, F.~Azhar, A.~Rodriguez, A.~Joulin, E.~Grave, and G.~Lample, ``Llama: Open and efficient foundation language models,'' 2023. [Online]. Available: \url{https://arxiv.org/abs/2302.13971}
\BIBentrySTDinterwordspacing

\bibitem{touvron2023llama2openfoundation}
\BIBentryALTinterwordspacing
Meta, ``Llama 2: Open foundation and fine-tuned chat models,'' 2023. [Online]. Available: \url{https://arxiv.org/abs/2307.09288}
\BIBentrySTDinterwordspacing

\bibitem{grattafiori2024llama3herdmodels}
\BIBentryALTinterwordspacing
------, ``The llama 3 herd of models,'' 2024. [Online]. Available: \url{https://arxiv.org/abs/2407.21783}
\BIBentrySTDinterwordspacing

\bibitem{10.1145/3696271.3696292}
\BIBentryALTinterwordspacing
J.~Zhang, Z.~Mai, Z.~Xu, and Z.~Xiao, ``Is llama 3 good at identifying emotion? a comprehensive study,'' in \emph{Proceedings of the 2024 7th International Conference on Machine Learning and Machine Intelligence (MLMI)}, ser. MLMI '24.\hskip 1em plus 0.5em minus 0.4em\relax New York, NY, USA: Association for Computing Machinery, 2024, p. 128–132. [Online]. Available: \url{https://doi.org/10.1145/3696271.3696292}
\BIBentrySTDinterwordspacing

\bibitem{10458651}
Z.~Xiao, Z.~Mai, Z.~Xu, Y.~Cui, and J.~Li, ``Corporate event predictions using large language models,'' in \emph{2023 10th International Conference on Soft Computing \& Machine Intelligence (ISCMI)}.\hskip 1em plus 0.5em minus 0.4em\relax IEEE, 2023, pp. 193--197.

\bibitem{xiao-etal-2024-analyzing}
\BIBentryALTinterwordspacing
Z.~Xiao, Y.~Huang, and E.~Blanco, ``Analyzing large language models{'} capability in location prediction,'' in \emph{Proceedings of the 2024 Joint International Conference on Computational Linguistics, Language Resources and Evaluation (LREC-COLING 2024)}, N.~Calzolari, M.-Y. Kan, V.~Hoste, A.~Lenci, S.~Sakti, and N.~Xue, Eds.\hskip 1em plus 0.5em minus 0.4em\relax Torino, Italia: ELRA and ICCL, May 2024, pp. 951--958. [Online]. Available: \url{https://aclanthology.org/2024.lrec-main.85}
\BIBentrySTDinterwordspacing

\bibitem{10.1145/3655497.3655500}
\BIBentryALTinterwordspacing
Z.~Xiao, Z.~Mai, Y.~Cui, Z.~Xu, and J.~Li, ``Short interest trend prediction with large language models,'' in \emph{Proceedings of the 2024 International Conference on Innovation in Artificial Intelligence}, ser. ICIAI '24.\hskip 1em plus 0.5em minus 0.4em\relax New York, NY, USA: Association for Computing Machinery, 2024, p.~1. [Online]. Available: \url{https://doi.org/10.1145/3655497.3655500}
\BIBentrySTDinterwordspacing

\bibitem{10.5555/3600270.3602070}
J.~Wei, X.~Wang, D.~Schuurmans, M.~Bosma, B.~Ichter, F.~Xia, E.~H. Chi, Q.~V. Le, and D.~Zhou, ``Chain-of-thought prompting elicits reasoning in large language models,'' in \emph{Proceedings of the 36th International Conference on Neural Information Processing Systems}, ser. NIPS '22.\hskip 1em plus 0.5em minus 0.4em\relax Red Hook, NY, USA: Curran Associates Inc., 2024.

\bibitem{zhang2022automaticchainthoughtprompting}
\BIBentryALTinterwordspacing
Z.~Zhang, A.~Zhang, M.~Li, and A.~Smola, ``Automatic chain of thought prompting in large language models,'' 2022. [Online]. Available: \url{https://arxiv.org/abs/2210.03493}
\BIBentrySTDinterwordspacing

\bibitem{wang2023selfconsistencyimproveschainthought}
\BIBentryALTinterwordspacing
X.~Wang, J.~Wei, D.~Schuurmans, Q.~Le, E.~Chi, S.~Narang, A.~Chowdhery, and D.~Zhou, ``Self-consistency improves chain of thought reasoning in language models,'' 2023. [Online]. Available: \url{https://arxiv.org/abs/2203.11171}
\BIBentrySTDinterwordspacing

\bibitem{zhao-etal-2024-enhancing-zero}
\BIBentryALTinterwordspacing
X.~Zhao, M.~Li, W.~Lu, C.~Weber, J.~H. Lee, K.~Chu, and S.~Wermter, ``Enhancing zero-shot chain-of-thought reasoning in large language models through logic,'' in \emph{Proceedings of the 2024 Joint International Conference on Computational Linguistics, Language Resources and Evaluation (LREC-COLING 2024)}, N.~Calzolari, M.-Y. Kan, V.~Hoste, A.~Lenci, S.~Sakti, and N.~Xue, Eds.\hskip 1em plus 0.5em minus 0.4em\relax Torino, Italia: ELRA and ICCL, May 2024, pp. 6144--6166. [Online]. Available: \url{https://aclanthology.org/2024.lrec-main.543}
\BIBentrySTDinterwordspacing

\bibitem{hu2024chainofsymbolpromptingelicitsplanning}
\BIBentryALTinterwordspacing
H.~Hu, H.~Lu, H.~Zhang, Y.-Z. Song, W.~Lam, and Y.~Zhang, ``Chain-of-symbol prompting elicits planning in large langauge models,'' 2024. [Online]. Available: \url{https://arxiv.org/abs/2305.10276}
\BIBentrySTDinterwordspacing

\bibitem{10.5555/3666122.3666639}
S.~Yao, D.~Yu, J.~Zhao, I.~Shafran, T.~L. Griffiths, Y.~Cao, and K.~Narasimhan, ``Tree of thoughts: deliberate problem solving with large language models,'' in \emph{Proceedings of the 37th International Conference on Neural Information Processing Systems}, ser. NIPS '23.\hskip 1em plus 0.5em minus 0.4em\relax Red Hook, NY, USA: Curran Associates Inc., 2024.

\bibitem{yao-etal-2024-got}
\BIBentryALTinterwordspacing
Y.~Yao, Z.~Li, and H.~Zhao, ``{G}o{T}: Effective graph-of-thought reasoning in language models,'' in \emph{Findings of the Association for Computational Linguistics: NAACL 2024}, K.~Duh, H.~Gomez, and S.~Bethard, Eds.\hskip 1em plus 0.5em minus 0.4em\relax Mexico City, Mexico: Association for Computational Linguistics, Jun. 2024, pp. 2901--2921. [Online]. Available: \url{https://aclanthology.org/2024.findings-naacl.183}
\BIBentrySTDinterwordspacing

\bibitem{zhou2023threadthoughtunravelingchaotic}
\BIBentryALTinterwordspacing
Y.~Zhou, X.~Geng, T.~Shen, C.~Tao, G.~Long, J.-G. Lou, and J.~Shen, ``Thread of thought unraveling chaotic contexts,'' 2023. [Online]. Available: \url{https://arxiv.org/abs/2311.08734}
\BIBentrySTDinterwordspacing

\bibitem{wang2024chainoftableevolvingtablesreasoning}
\BIBentryALTinterwordspacing
Z.~Wang, H.~Zhang, C.-L. Li, J.~M. Eisenschlos, V.~Perot, Z.~Wang, L.~Miculicich, Y.~Fujii, J.~Shang, C.-Y. Lee, and T.~Pfister, ``Chain-of-table: Evolving tables in the reasoning chain for table understanding,'' 2024. [Online]. Available: \url{https://arxiv.org/abs/2401.04398}
\BIBentrySTDinterwordspacing

\bibitem{10871796}
Y.~Wu, Z.~Xiao, J.~Zhang, Z.~Mai, and Z.~Xu, ``Can llama 3 understand monetary policy?'' in \emph{2024 17th International Conference on Advanced Computer Theory and Engineering (ICACTE)}, 2024, pp. 145--149.

\bibitem{10.1145/3637528.3671647}
\BIBentryALTinterwordspacing
M.~Wan, T.~Safavi, S.~K. Jauhar, Y.~Kim, S.~Counts, J.~Neville, S.~Suri, C.~Shah, R.~W. White, L.~Yang, R.~Andersen, G.~Buscher, D.~Joshi, and N.~Rangan, ``Tnt-llm: Text mining at scale with large language models,'' in \emph{Proceedings of the 30th ACM SIGKDD Conference on Knowledge Discovery and Data Mining}, ser. KDD '24.\hskip 1em plus 0.5em minus 0.4em\relax New York, NY, USA: Association for Computing Machinery, 2024, p. 5836–5847. [Online]. Available: \url{https://doi.org/10.1145/3637528.3671647}
\BIBentrySTDinterwordspacing

\bibitem{javadi2024llmbasedweaksupervisionframework}
\BIBentryALTinterwordspacing
F.~Javadi, P.~Gampa, A.~Woo, X.~Geng, H.~Zhang, J.~Sepulveda, B.~Bayar, and F.~Wang, ``Llm-based weak supervision framework for query intent classification in video search,'' 2024. [Online]. Available: \url{https://arxiv.org/abs/2409.08931}
\BIBentrySTDinterwordspacing

\bibitem{10.1145/3696271.3696294}
\BIBentryALTinterwordspacing
Z.~Mai, J.~Zhang, Z.~Xu, and Z.~Xiao, ``Is llama 3 good at sarcasm detection? a comprehensive study,'' in \emph{Proceedings of the 2024 7th International Conference on Machine Learning and Machine Intelligence (MLMI)}, ser. MLMI '24.\hskip 1em plus 0.5em minus 0.4em\relax New York, NY, USA: Association for Computing Machinery, 2024, p. 141–145. [Online]. Available: \url{https://doi.org/10.1145/3696271.3696294}
\BIBentrySTDinterwordspacing

\bibitem{zhang2024empowering}
J.~Zhang, K.~Lu, Y.~Wan, J.~Xie, and S.~Fu, ``Empowering uav-based airborne computing platform with sdr: Building an lte base station for enhanced aerial connectivity,'' \emph{IEEE Transactions on Vehicular Technology}, 2024.

\bibitem{URLLC}
\BIBentryALTinterwordspacing
3GPP, ``{Reliable and Low Latency Communication},'' 2023, {A}ccessed: 2024-12-23. [Online]. Available: \url{https://www.3gpp.org/technologies/urlcc-2022}
\BIBentrySTDinterwordspacing

\end{thebibliography}
\bibliographystyle{IEEEtran}

\end{document}